\documentclass[letterpaper, 10 pt, conference]{ieeeconf}  

\IEEEoverridecommandlockouts                              

\title{\LARGE \bf
Interactive Navigation in Environments with Traversable Obstacles Using Large Language and Vision-Language Models
}

\author{Zhen Zhang$^{1,2}$, Anran Lin$^{3}$, Chun Wai Wong$^{1}$, Xiangyu Chu$^{1,2,4}$, Qi Dou$^{3}$, and K. W. Samuel Au$^{1,2}$
\thanks{This work was supported in part by Chow Yuk Ho Technology Centre of Innovative Medicine, The Chinese University of Hong Kong, in part by the Multiscale Medical Robotics Centre, AIR@InnoHK, and in part by the Research Grants Council (RGC) of Hong Kong under Grant 14211320. 
$^{1}$: Department of Mechanical and Automation Engineering, The Chinese University of Hong Kong, Hong Kong SAR.  $^{2}$: Multiscale Medical Robotics Centre, Hong Kong SAR. $^{3}$: Department of Computer Science and Engineering, The Chinese University of Hong Kong, Hong Kong SAR. $^{4}$: Computer Aided Medical Procedures, Technical University of Munich, Germany.  (\textit{Corresponding authors: Xiangyu Chu and Qi Dou})}%
}

\usepackage{graphicx}
\usepackage{subfigure} 
\usepackage{float}
\usepackage{mathtools}
\usepackage{cite}
\usepackage{xcolor}
\usepackage{tcolorbox}
\usepackage[ruled,linesnumbered]{algorithm2e}
\usepackage{amsmath,amssymb,mathrsfs}
\usepackage{comment}
\usepackage{url}
\usepackage{pifont}
\usepackage[]{hyperref}
\begin{document}

\maketitle
\thispagestyle{empty}
\pagestyle{empty}

\begin{abstract}


This paper proposes an interactive navigation framework by using large language and vision-language models, allowing robots to navigate in environments with traversable obstacles. We utilize the large language model (GPT-3.5) and the open-set Vision-language Model (Grounding DINO) to create an action-aware costmap to perform effective path planning without fine-tuning.  With the large models, we can achieve an end-to-end system from textual instructions like ``Can you pass through the curtains to deliver medicines to me?", to bounding boxes (e.g., curtains) with action-aware attributes. They can be used to segment LiDAR point clouds into two parts: traversable and untraversable parts, and then an action-aware costmap is constructed for generating a feasible path. The pre-trained large models have great generalization ability and do not require additional annotated data for training, allowing fast deployment in the interactive navigation tasks. We choose to use multiple traversable objects such as curtains and grasses for verification by instructing the robot to traverse them. Besides, traversing curtains in a medical scenario was tested. All experimental results demonstrated the proposed framework's effectiveness and adaptability to diverse environments.  


\end{abstract}

\section{INTRODUCTION}
\begin{figure}[htbp]
    \centering
    \includegraphics[width=0.35\textwidth]{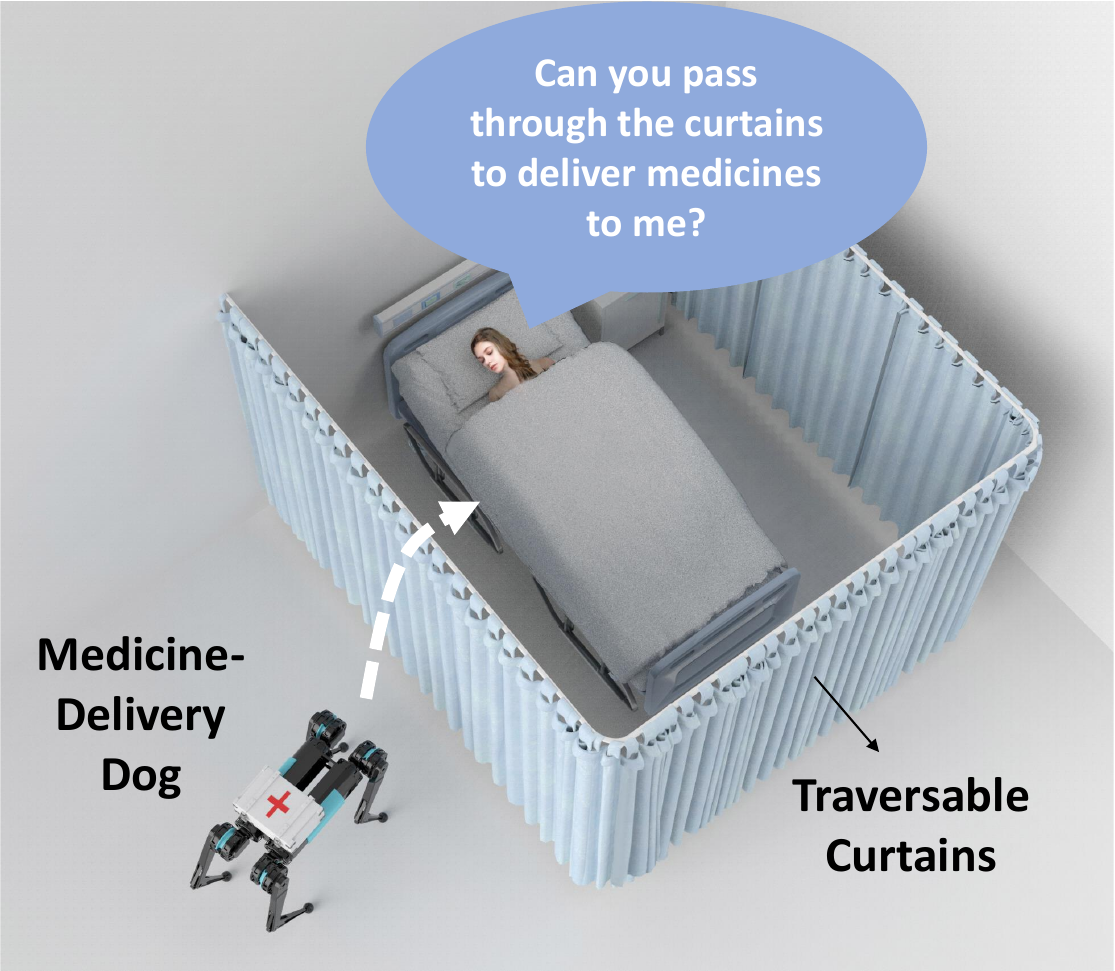}
    \caption{An example of interactive navigation. When a medicine-delivery dog arrives at a room, a patient can further interact with the robotic dog and ask it to walk to the bed by passing through the traversable curtains. Such an interaction can help navigate the robot to the place where it cannot normally reach and therefore meet humans' real-time needs.}
\label{Illustration_interactive_navigation}
\end{figure}

Robot navigation is critical in deploying robots in an obstacle-involved workspace. Besides the long-sought need for safety and efficiency~\cite{wellhausen2020safe,vulcano2022safe, guan2022ga}, interaction with humans to satisfy real-time needs is becoming more appealing. For example, a patient can interact with a medicine-delivery dog and ask it to enter the ward area by passing through the surrounding curtains, as shown in Fig.~\ref{Illustration_interactive_navigation}. Without real-time interaction and extra treatments, it is challenging for the robotic dog to pass through the curtain because the curtain is normally considered as an obstacle in its motion planner based on its sensor data (e.g., LiDAR) and thus no feasible path can be found. Recently, large language models have provided an alternative solution to real-time human-robot interaction, but using them to interact with robots for navigation in various environments remains open. In this paper, we focus on large-model-based interactive navigation for robots, especially in environments with traversable obstacles like curtains and grasses for a robotic dog. 

\begin{figure*}[htbp]
    \centering
    \includegraphics[width=0.95\textwidth]{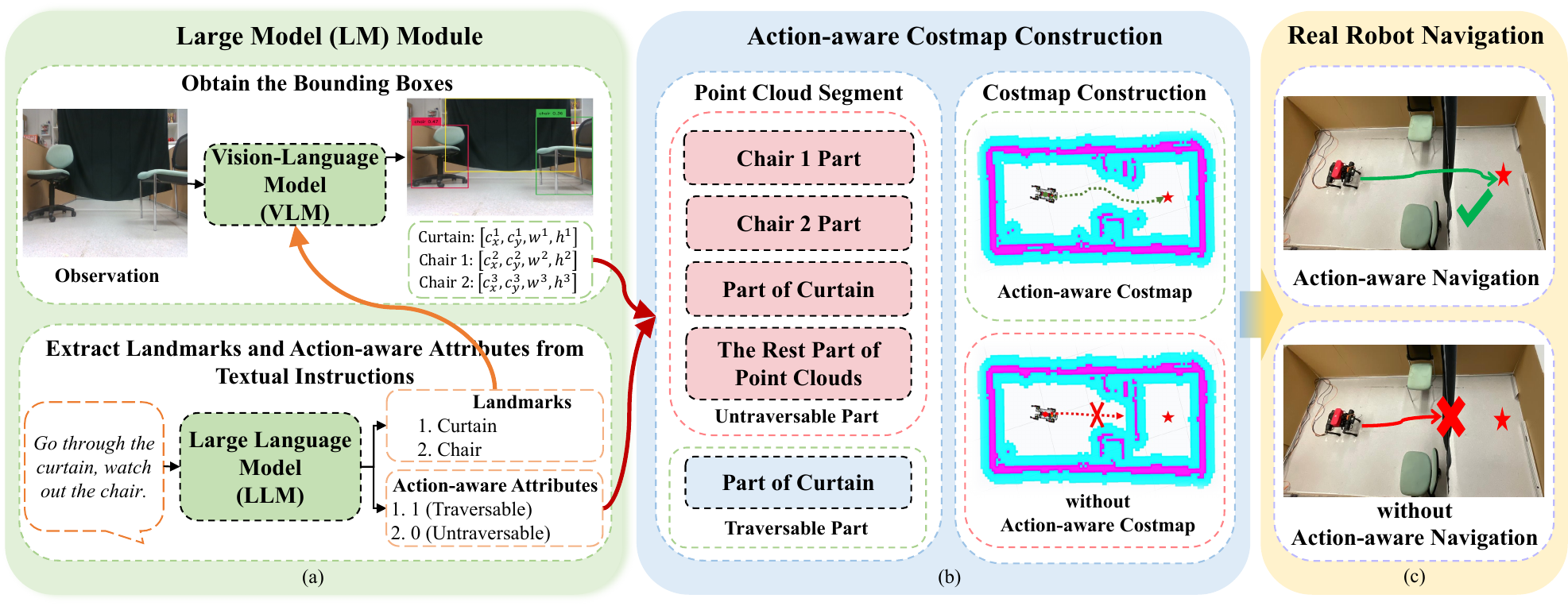}
    \caption{Proposed large-model-based interactive navigation framework for the robots in environments with traversable obstacles. (a) The large model module uses LLM and VLM to extract the landmarks' bounding boxes and action-aware attributes from texts/speeches. (b) The action-aware costmap is constructed with the segmented point clouds based on the output of the large model module. (c) A feasible path can be planned with an action-aware costmap, while no feasible path would be found if not considering the landmarks' action-aware attribute.}
    \label{navigation_framework}
\end{figure*}

Interacting with robots for navigation has been studied via many interfaces such as gestures, teleoperation devices, and natural languages.~\cite{liu2019obstacle, naseer2013followme, ali2012real} recognized simple human gestures by using onboard cameras to control robots' behavior like their forward-backward direction. Besides, different teleoperation devices such as joysticks~\cite{stotko2019vr} and tablets~\cite{olatunji2022levels} can be used for navigating robots to desired places. We find these two kinds of interfaces may not be available for patients in bed. A more natural interface is to use voices, i.e., talking to the robot. Early works on robot navigation using natural languages relied on natural language processing techniques to obtain semantic information such as landmarks~\cite{hu2019safe}. Recently, the birth of Large Language Models (LLMs) greatly improved the capability of extracting semantic information~\cite{brown2020language,taylor2022galactica,hoffmann2022training,wei2022chain}. Although LLMs have shown much better generalization, their output cannot be understood by navigation algorithms. For example, a motion planner cannot understand ``curtains" extracted by LLMs. To bridge the gap, people started to combine language and vision-language large models.~\cite{shah2023lm} used Vision-and-Language Models (VLMs) to associate images with texts from LLMs and then a visual navigation model is used to generate a feasible path based on the images from VLMs. Since the combination of a pre-trained language and a vision-language model can parse complex high-level instructions for robots, in this paper, we choose to use this paradigm to deal with human's natural language in real-time interaction. 
In our case, we cannot use visual navigation methods to generate feasible paths because the landmarks in visual navigation methods (e.g.,~\cite{shah2023lm, shah2023vint, anderson2018vision, gasparino2022wayfast}) are normally assumed to be untraversable, thus no feasible path can be found when the goal (e.g., a bed) cannot be detected (see Fig.~\ref{Illustration_interactive_navigation}).

Constructing costmaps is an alternative solution to generate a feasible path. There are some works on connecting images with costmaps. For example,~\cite{hu2019safe} made instance segmentation to extract semantic objects of interest in the current image. Then the obtained masks are fed to the depth image for objects' localization, and finally, the costmap would be updated to include the objects like a moving human.~\cite{huang2023visual} also built different obstacle maps for different embodiments given images. For example, a table is an obstacle for a wheeled robot but not for a drone. Compared to visual navigation that relies on goal images, constructing costmaps is more fundamental, and shifts the feasible path generation to the navigation planner. However, constructing costmaps based on the landmark images is not sufficient in our task because the landmarks are not traversable in default and thus the robot would get stuck in front of the landmarks.
Unlike~\cite{shah2023lm} where only landmarks were extracted and verbs/commands were disregarded, we propose to extract action-aware attributes corresponding to the landmarks and then use them for costmap construction. For example, ``pass through the curtains" means the curtains are traversable due to ``pass through". Such an action-aware property integrated into building the costmaps allows the robot to pass through the curtains.

In this paper, we propose an action-aware interactive navigation system based on pre-trained large models(i.e., LLM and VLM). LLM is used to extract landmarks and corresponding action-aware attributes in texts, and VLM is used to detect the landmarks in the picture and obtain the corresponding bounding boxes. These large models help to make the transformation from speeches and texts to bounding boxes and their action-aware attributes without any fine-tuning and additional training data. 
\cite{chen20232} used large models to classify different navigation subtasks and required multiple action-specific navigators. Different from \cite{chen20232}, we use the action-aware attributes of landmarks obtained from large models to fundamentally build an action-aware costmap for generating a feasible path and thus achieve more intelligent and flexible interactive navigation that meets humans' real-time needs. We summarize the contributions of this paper:\\
1) We propose a large-model-based interactive navigation framework for robots to plan a feasible path in environments with traversable objects.\\
2) Besides landmarks, action-aware attributes in textual instructions are extracted to assist sensor data segmentation. Such an action-aware attribute allows us to construct an action-aware costmap for interactive navigation.\\
3) Extensive experiments demonstrate the effectiveness of the proposed framework and its generalization on different traversable objects and scenarios.




\section{PROPOSED FRAMEWORK}

\subsection{Problem Formulation}

Given a set of landmarks and their corresponding action-aware attributes $\{ (\ell_1, P_{a_{1}|\ell_{1}}), (\ell_2, P_{a_{2}|\ell_{2}}), \cdots, (\ell_n, P_{a_{n}|\ell_{n}})\}$ extracted by LLM from language-based interaction. $P_{a_{i}|\ell_{i}} = 1$ denotes that the landmark $\ell_i$ associated with the action $a_i$ is traversable and $P_{a_{i}|\ell_{i}} = 0$ denotes that the landmark $\ell_i$ associated with the action $a_i$ is untraversable. With a goal in the static map, our objective is to find a feasible path including a series of waypoints $\{ w_1, w_2, \cdots, w_m\}$. In this path, all untraversable landmarks' action-aware attributes are respected (i.e., maintaining a safe distance from these landmarks) and traversable landmarks are fully utilized to achieve the goal. 

Note that finding a feasible path is not a trivial problem when we consider landmarks' action-aware attributes. For example, in Fig.~\ref{Illustration_interactive_navigation}, the bed cannot be detected if using visual sensors before traversing the curtains and entering the ward area, thus no feasible path can be found. Without human interaction, the curtains are still regarded as obstacles and it takes more effort to get close to the bed. 

\subsection{Framework Overview}
To solve the problem, we propose a large-model-based interactive navigation framework that can map language commands to a feasible path, as shown in Fig.~\ref{navigation_framework}. This framework consists of three parts: a large model module, action-aware costmap construction, and a navigation planner. 

The large model module includes a LLM and a VLM. LLMs can learn the grammar and semantics of natural language by learning massive corpus data. In this work, the pre-trained LLM does not require additional data for training and fine-tuning. It can be used to extract landmarks $\ell_i$ and their corresponding action-aware attributes $P_{a_{i}|\ell_{i}}$ in the text. 
With VLMs, we can detect objects in images using language~\cite{chen2021pix2seq,du2022learning,li2022grounded,liu2023grounding}. The key is to associate the semantic information of the language with the image feature information of the object. The output of the VLM is the information on the bounding boxes of the landmarks that occur in the language instructions.

The costmap construction part utilizes the output of the large model module, including the landmarks' bounding boxes and action-aware attributes, to build up an action-aware costmap. The costmap is a container that can store information about the environment and can be further used for path planning~\cite{5477164,5980048}. Moreover, the layered costmap~\cite{lu2014layered} can include multiple costmap layers according to different sensor data and semantic information, which can solve the path planning problem in complex scenarios\cite{7324184,7140063,7745154}.
In the proposed framework, we construct such a flexible layered costmap with semantic information related to the landmarks in the form of bounding boxes and their action-aware attributes obtained from the large model module.

With an action-aware costmap, many off-the-shelf path planning methods can be employed for the part of robot navigation. In this paper, we choose to use the A star ($A^{*}$) search algorithm~\cite{hart1968formal}. A feasible path including a series of waypoints $\{ w_1, w_2, \cdots, w_m\}$ can be obtained, which is not attainable without action-aware costmaps (see Fig.~\ref{navigation_framework} (b)).

\section{LARGE MODEL DEPLOYMENT}
In this work, we use GPT-3.5 as the LLM to extract the landmarks and corresponding action-aware attributes in textual instructions which can be transferred from speeches, and then use the VLM (Grounding-DINO~\cite{liu2023grounding}) to locate the bounding boxes corresponding to the landmarks, including a center point coordinate $(c_{x}, c_{y})$, width $w$, and height $h$ in pixel for each bounding box, denoted as $B = \{c_x, c_y, w, h \}$. 

\subsection{Large Language Model}
We deploy GPT-3.5 by accessing OpenAI’s API\footnote{https://openai.com/api/}. To perform interactive navigation, we introduce an \textbf{action-aware} attribute corresponding to a landmark. We define that objects in the real-world environment can have two kinds of action-aware attributes. One is ``traversable", which means a robot can traverse the landmark. Another is ``untraversable", and a robot cannot traverse the landmark as much previous work assumed. 

To describe the action-aware attribute of a landmark $\ell$, we define:

\begin{equation}
P_{a|\ell}=
\begin{cases}
1  & \rightarrow \text{$\ell$ is traversable under $a$} \\
0 & \rightarrow \text{$\ell$ is untraversable under $a$}
\end{cases},
\end{equation}
where $a$ is an action (e.g., ``go through" and ``traverse") related to the landmark.




With GPT-3.5, we can extract landmarks and action-aware attributes from free-form instructions. The large model is robust to different descriptions in the input prompts and is also greatly effective in parsing instructions. In the following, we make prompt engineering and then show a test example.

\begin{center}
\begin{tcolorbox}[left = 1mm, right = 1mm, top = 1mm, bottom = 1mm, colback=gray!10, 
                  colframe=black, 
                  width=0.48\textwidth,
                  arc=1mm, auto outer arc,
                  boxrule=0.5pt,
                 ] \small
\textit{\textbf{Example 1}: I ask: ``Please go through the curtain and watch out for the medicine trolley". You should respond ``curtain" and ``1"; ``trolley" and ``0".}


\vspace*{6pt}
\textit{\textbf{Example 2}: I ask: ``Please pass through the curtain but be careful of the table in the middle of the room". You should respond ``curtain" and ``1"; ``table" and ``0".}
\end{tcolorbox}
\end{center}



Here is an example. The test textual instruction is: ``\textit{Go through the curtain, and watch out the chair}.". The test output is: \textit{curtain and 1; chair and 0}.

\subsection{Vision-Language Model}
The landmarks obtained from the LLM output are fed to the VLM in the form of a simple prompt \textit{``curtain, chair"}. 
We utilize Grounding DINO \cite{liu2023grounding} to obtain the bounding box $B$ for each landmark. 
Specifically, for each \textit{(Image, Text)} pair, image and text features are extracted by an image backbone and a text backbone respectively, and then fed into a feature enhancer module to fuse cross-modality features. After that, cross-modality queries are selected using a language-guided query selection module and fed into a cross-modality decoder. The output queries of the last decoder layer are used to predict object boxes $B = \{c_x, c_y, w, h \}$ and extract corresponding phrases (e.g., \textit{curtain}). 
\begin{figure}[htbp]
    \centering
    \includegraphics[width=0.48\textwidth]{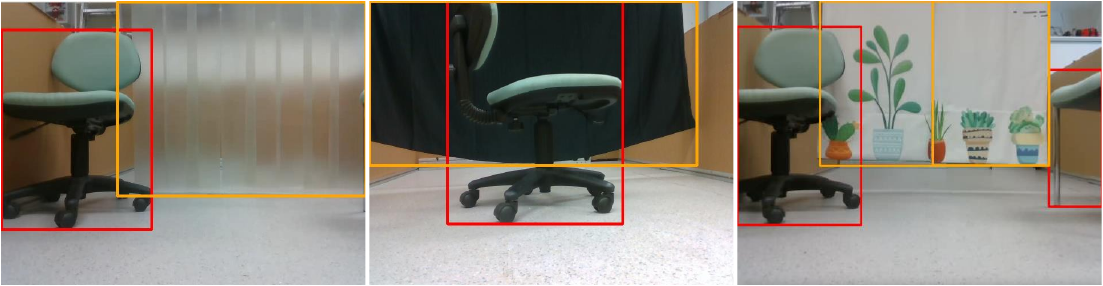}
    \caption{Examples of output of the large model module. The initial instruction for these examples is: “Please go through the curtain and be careful of the chair.”. We obtain the landmarks and the corresponding traversable attributes, \textit{curtain and 1} (in orange); \textit{chair and 0} (in red), using GPT-3.5. Then, we can ground the related bounding boxes with~\cite{liu2023grounding}.}
    \label{LLM_result}
\end{figure}
Therefore, the final outputs of the large model module for each landmark form the input, $\{c_x, c_y, w, h, P_{a|\ell}\}$, for the following action-aware costmap construction. Figure \ref{LLM_result} shows some examples. It should be noted that in the example in the lower right corner, due to the limitation of the VLM, a certain part of the object may be output as a bounding box (see the two bounding boxes of the curtain). With the help of the method in Sec.~\ref{LiDAR point segmentation}, we can also handle this situation well.



\section{ACTION-AWARE COSTMAP CONSTRUCTION}
With LLM and VLM, we obtain the 2D bounding boxes $ \{c_{x},c_{y},w,h\}$ of the landmarks in the picture and the action-aware attribute $P_{a|\ell}$ corresponding to each landmark $\ell$. 
To find a feasible path, we propose an action-aware costmap construction method. In our system, we use a LiDAR mapping sensor. First, We transform the LiDAR point clouds to the image plane.
Then, according to the bounding boxes and their corresponding action-aware attributes obtained by the large model module, the LiDAR point clouds are segmented into two parts: the traversable part and the untraversable part. Lastly, we use the two parts of LiDAR point clouds as input sensor sources to build an action-aware costmap and update the map in real-time. The costmap construction pipeline accompanied by the processing based on large models can be found in Algorithm~\ref{algorithm}.

\begin{algorithm}[h]\footnotesize

\caption{Action-aware Costmap Construction with Large Language and Vision-Language Models}
\label{algorithm}
 \SetKwData{Left}{left}\SetKwData{This}{this}\SetKwData{Up}{up}
  \SetKwFunction{Union}{Union}\SetKwFunction{FindCompress}{FindCompress}
  \SetKwInOut{Input}{Input}\SetKwInOut{Output}{Output}
\Input{Textual instructions $T_{t}$ or Voice $V_{t}$, camera image $I_{t}$, current LiDAR point clouds $PC_{t}$. $t$ is the current step.}
\Output{An action-aware costmap $M_{t}$}

·eIf{$V_{t}$ is not None}{$T_{t}$ = Speech\_recognition($V_{t}$)\;}
{$T_{t}$ = $T_{t}$\;}

$\left\{\left(\ell_{1}, P_{a_{1}|\ell_{1}}\right), \left(\ell_{2}, P_{a_{2}|\ell_{2}}\right),...,  \left(\ell_{n}, P_{a_{n}|\ell_{n}}\right)
\right\}$ = $LLM(T_{t})$\;



\While{$\Vert current\_position - target \Vert_{2} > Threshold$}
{
$\left\{B_{1},B_{2},...,B_{m}\right\}$ = \textit{VLM}$\left(I_{t},\left(\ell_{1},\ell_{2},...,\ell_{n}\right)\right)$, \text{where} $B_{i} = \{c_{x}^{i},c_{y}^{i},w^{i},h^{i}\}$\;
$P_{\mathcal{A}_{i}} = P_{a_{i}|\ell_{i}}$, where $\mathcal{A}_{i}= \{c_x^{i}-\frac{w}{2}\leq x \leq c_x^{i}+\frac{w}{2}, c_y^{i}-\frac{h}{2}\leq y \leq c_y^{i}+\frac{h}{2}\}$\;

\eIf{$\left\{\mathcal{A}_{1},\mathcal{A}_{2},...,\mathcal{A}_{m}\right\}$ is not empty}
{
$\left[PC_{t}^{Tra},PC_{t}^{Untra}\right]$ = 
\qquad  Segment$\left(PC_{t},\left\{(\mathcal{A}_1, P_{\mathcal{A}_1}),...,(\mathcal{A}_m, P_{\mathcal{A}_m})\right\}\right)$\;
$M_{t}$ = UpdateCostmap$\left(M_{t-1},\left(PC_{t}^{Tra},PC_{t}^{Untra}\right)\right)$;}
{
$M_{t}$ = UpdateCostmap$\left(M_{t-1}, PC_{t}\right)$\;
}

}

\end{algorithm}

\subsection{Transformation from LiDAR to Camera}
To segment the LiDAR point clouds with the obtained bounding boxes, the LiDAR point clouds need to be transformed into the image plane. We can use the extrinsic matrix and intrinsic matrix to project a 3D point cloud onto the 2D image plane of the camera. Given a LiDAR point $p =(x_{i}, y_{i}, z_{i})$ in the Cartesian coordinate system with $z$-axis pointing upward, it can be transformed into the image plane and obtain a coordinate $s = (u, v)$~\cite{hartley_zisserman_2004}:
\begin{equation}
        \begin{bmatrix}u& v&1\end{bmatrix}^{T}  =
\begin{bmatrix}
    \frac{u^{'}}{w^{'}}&\frac{v^{'}}{w^{'}}&1
\end{bmatrix}^{T},
\end{equation}
$$
\begin{bmatrix}
    u^{'}\\v^{'}\\w^{'}
\end{bmatrix}  = \underbrace{
        \begin{bmatrix}
          fs_{x} & ks_{y} & u_{0} \\
          0 & fs_{y} & v_{0} \\
          0 & 0 & 1
         \end{bmatrix}}_{intrinsic}
    \begin{bmatrix}
    \mathbf{I}_{3\times 3}\\\mathbf{0}_{1\times 3} \end{bmatrix}^{T} 
     \underbrace{
    \begin{bmatrix}
     R_{L}^{C} & t_{L}^{C} \\
     \mathbf{0}_{1\times 3} & 1 
    \end{bmatrix}}_{extrinsic}
\begin{bmatrix} x_{i} \\ y_{i} \\ z_{i} \\1 \end{bmatrix}, $$
where $f$ is the focal length, $s_{x}$ and $s_{y}$ are the scale in $x$ and $y$ respectively, $k
$ is the rotation of shear in the $y$ direction to than in $x$, $R_{L}^{C}$ and $t_{L}^{C}$ are the rotation matrix and translation vector from LiDAR frame to camera frame, $\left(u_{0}, v_{0}\right)$ is the distance from the image center to the image plane coordinate system origin, and $w^{'}$ is the distance from camera centre to the LiDAR point along $z$-axis. For simplification, we can define such a transformation as $s = \mathbb{T}(p)$. 
\begin{figure}[htbp]
    \centering
    \includegraphics[width=0.48\textwidth]{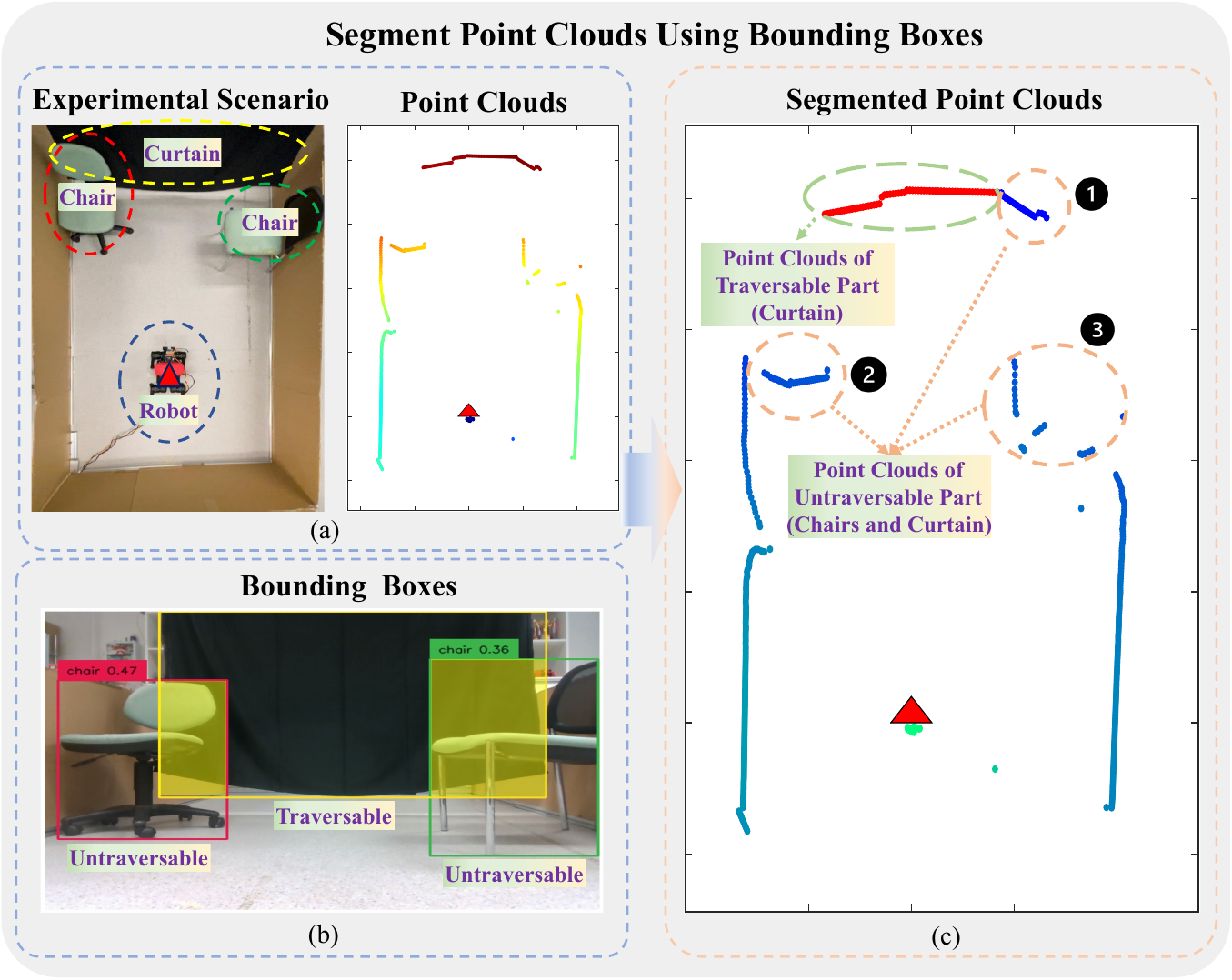}
    \caption{(a) The experimental scenario in bird's eye view and the point clouds obtained from LiDAR Mapping sensor. (b) The yellow areas are the intersection between the bounding boxes of untraversable and traversable objects. (c) The point clouds colored red are the traversable part belonging to the curtain. The rest is the untraversable part belonging to part of the curtain (\ding{202}), chairs (\ding{203}, \ding{204}), and the environment.}
    \label{action_aware_seg}
\end{figure}
\subsection{LiDAR Point Clouds Segmentation}\label{LiDAR point segmentation}
After projecting the LiDAR point clouds $PC_t$ into the image plane $I_t$, we can determine the action-aware attribute for each LiDAR point by checking if $s  = \mathbb{T}(p) \in \mathcal{A}$, where $\mathcal{A} = \text{Area}\{c_x-\frac{w}{2}\leq u \leq c_x+\frac{w}{2}, c_y-\frac{h}{2}\leq v \leq c_y+\frac{h}{2}\}$. Specifically, we create the masks with the same size as the bounding boxes and then use them to segment the LiDAR point clouds belonging to different landmarks on the image plane.  In practice, there are multiple overlapping bounding boxes, so one LiDAR point may belong to multiple bounding boxes. Based on the action-aware attributes of the landmarks $P_{a|\ell}$, the action-aware attribute for each LiDAR point in the bounding box is known, i.e., $P_\mathcal{A} = P_{a|\ell}$. Thus, a function \texttt{Segment} is defined as
\begin{equation}
p \in
\begin{cases}
PC_t^{Tra},  &   P_{\mathcal{A}_i} =1 ~ \text{if} ~ s \in \mathcal{A}_i, \forall i, \\
PC_t^{Untra}, & \text{else}
\end{cases},\\ \forall p \in PC_t, 
\end{equation}
where $i = 1,..., m$ and $m$ is the number of bounding boxes. A LiDAR point belonging to the traversable part $PC_t^{Tra}$ requires that all action-aware attributes of its occupied bounding boxes should be 1. When the bounding box with the traversable attribute overlaps with the bounding box with the untraversable attribute (see yellow intersection areas in Fig.~\ref{action_aware_seg} (b)), the overlapping part needs to be considered as the untraversable part $PC_t^{Untra}$ for safety. From Fig.~\ref{action_aware_seg} (c), it is observed that LiDAR point clouds are segmented into two parts (blue for untraversable and red for traversable) using bounding boxes with action-aware attributes. Especially, the circle 1 (\ding{202}) in Fig.~\ref{action_aware_seg} (c) is untraversable, although it is part of the traversable curtain.


\subsection{Costmap Building}
To build the costmap, we use the two parts of the LiDAR point clouds: $PC_t^{Tra}$ and $PC_t^{Untra}$. 
For the LiDAR points in $PC_t^{Tra}$, the cost on the costmap is 0, which is considered to be \textit{FREESPACE}, while for the LiDAR points in $PC_t^{Untra}$, its cost on the costmap is 254, which is considered to be \textit{LETHALSPACE}. With an action-aware costmap, a feasible path can be found since the cost of the LiDAR points related to the curtain has been set to 0. Otherwise, no feasible path can be obtained, see the costmap construction in Fig.~\ref{navigation_framework} (b).


\section{EVALUATION}
We tested the performance of our interactive navigation framework on the Unitree A1 quadruped robot platform equipped with a 2D LiDAR mapping sensor (\textit{Slamtec Mapper M2M2}) and a depth camera (\textit{Intel RealSense D435i}). The legged robot control used an open-source package~\cite{legged_control}. We ran the interactive navigation framework on a PC equipped with an Intel Core i7-13700K CPU. We demonstrated the results of the entire interactive navigation system in both simulated and real-world scenarios. Especially, in real-world scenarios, we conducted extensive experiments on curtains of different materials, types, and shapes to test the reliability and adaptability of the proposed framework. We also evaluated the performance of other objects with traversable properties such as grasses.



\subsection{Framework Evaluation in a Simulated Environment}

\begin{figure}[htbp]
    \centering
    \includegraphics[width=0.48\textwidth]{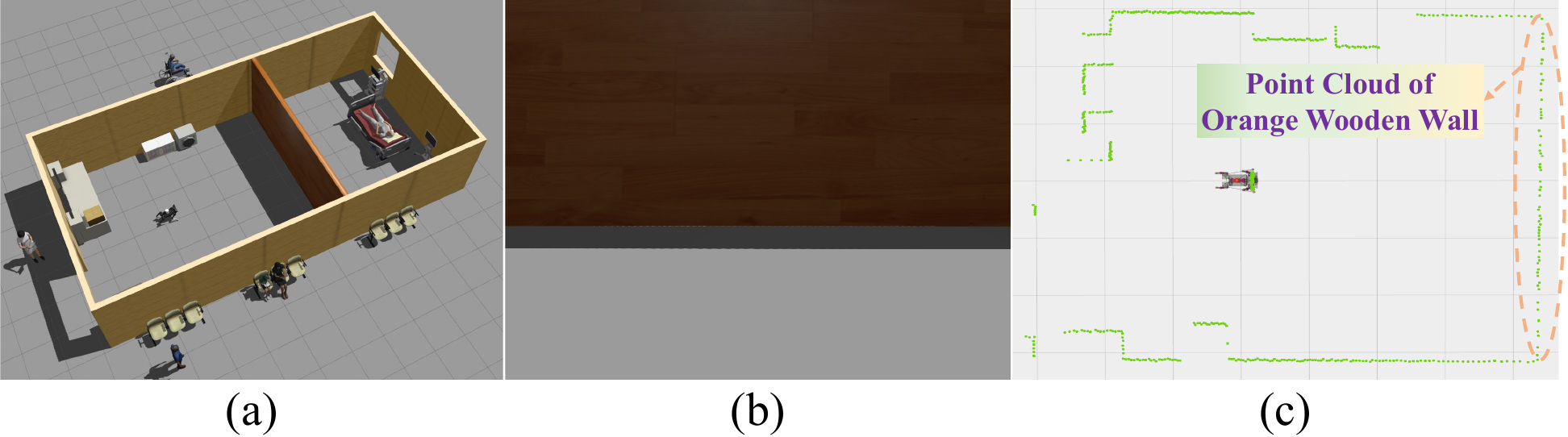}
    \caption{(a) We test our interactive navigation system in a simulated in-door hospital ward environment with static obstacles. (b-c) The simulated legged robot can obtain the RGB image and point clouds, respectively.   }
    \label{simulation_scene}
\end{figure}
As shown in Fig.~\ref{simulation_scene}, we used an indoor hospital scene in the Gazebo simulator~\cite{koenig2004design} to verify the feasibility of the framework. It contains static obstacles in the room such as ``elder lady patient", ``medical instrument", ``walker", ``orange wooden wall", ``freezer condenser", ``cabinet", and ``white table". To be noticed, the ``orange wooden wall" separates the hospital ward into two spaces.

\begin{figure}[htbp]
    \centering
    \includegraphics[width=0.48\textwidth]{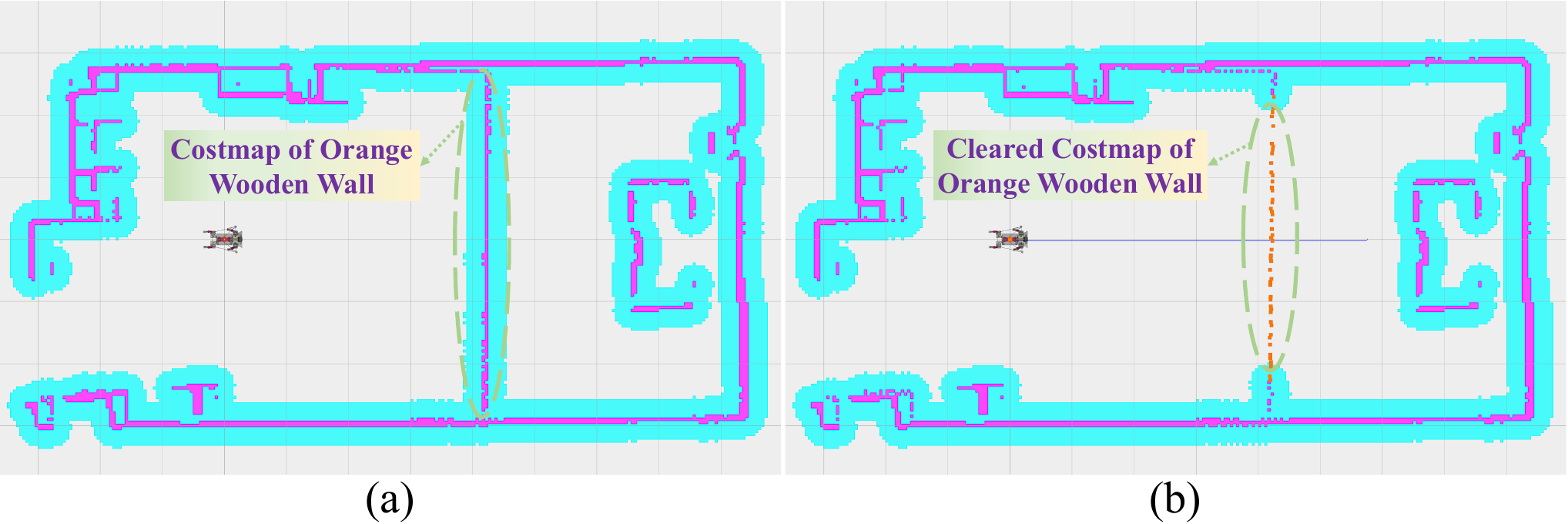}
    \caption{(a) Without using the action-aware costmap, the orange wooden wall would be regarded as an untraversable obstacle and marked in the costmap. Thus, no feasible path is planned. (b) A feasible path can be planned by using the action-aware costmap because part of the orange wooden wall is regarded as a traversable obstacle and cleared in the costmap.}
    \label{path_planning}
\end{figure}

In the simulation, we assumed that the action-aware attribute of the ``orange wooden wall" is traversable and our textual instruction can be ``Go through the orange wooden wall". Given a goal behind the wall, no feasible path can be planned without the action-aware costmap, as shown in Fig.~\ref{path_planning} (a). However, we found part of the ``orange wooden wall" can be cleared in the costmap if constructing an action-aware costmap and a straightforward path that crosses ``orange wooden wall" can be obtained, as shown in Fig.~\ref{path_planning} (b), demonstrating the effectiveness of the proposed framework.


\subsection{Framework Evaluation in Real-world Environments}
\begin{figure*}[htbp]
    \centering
   \includegraphics[width=0.96\textwidth]{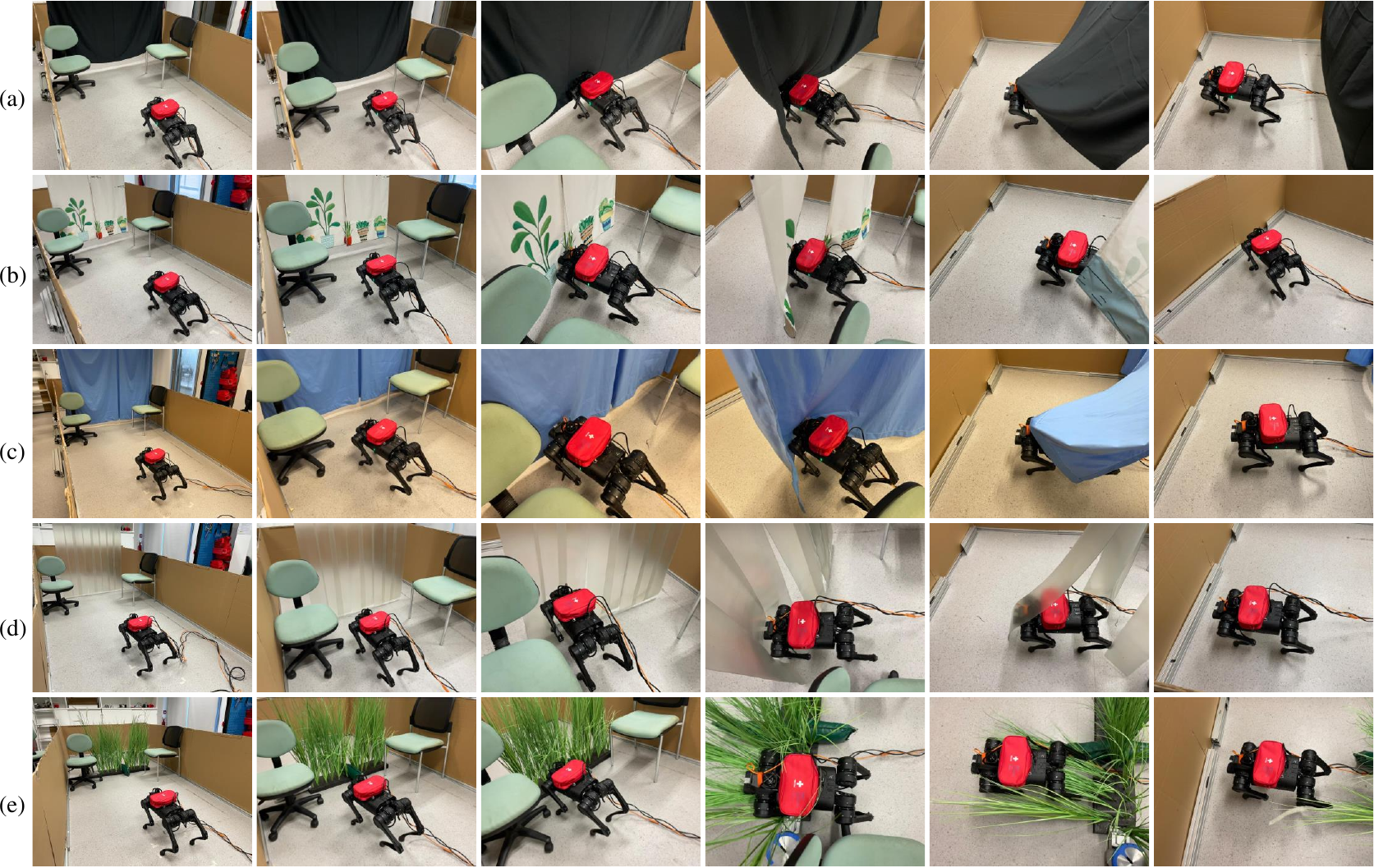}
    \caption{(a-d) Experiments of going through different curtains. (e) Experiment of going through the grass.}
    \label{lab_exp}
\end{figure*}

\begin{figure*}[htbp]
    \centering
   \includegraphics[width=0.96\textwidth]{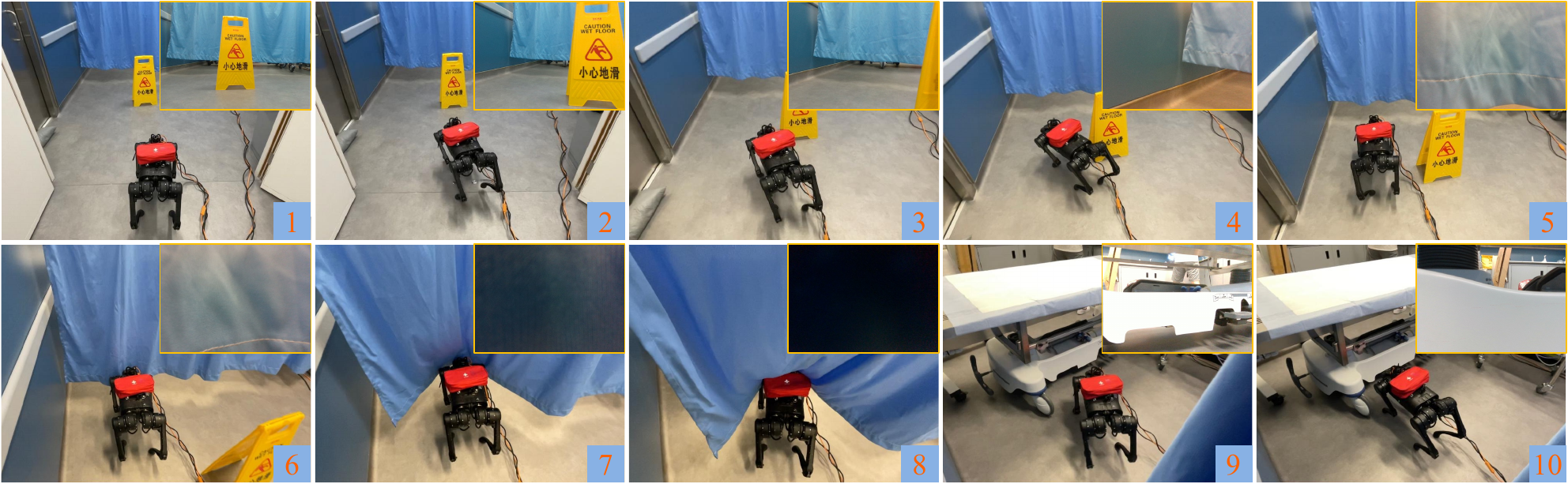}
    \caption{Experiment of the medical scenario at MRC. The upper right corner of each picture shows the first-person point of view of the robot.}
    \label{mrc_exp}
\end{figure*}
We conducted experiments in two real-world scenarios: a laboratory scenario and a medical scenario at the Multi-Scale Medical Robotics Center (MRC:https: //www.mrc-cuhk.com). We set up a simulation-like scenario in both and ran the SLAM algorithm~\cite{hess2016real} to construct a global map and annotate the goal (e.g., a bed in the scenario of MRC) in the map. For the traversable objects, We prepared 4 curtains with different colors and textures, which are common in our daily life, and we also prepared two different types of artificial grass. 
\subsubsection{Go through the curtain}
First, we conducted experiments with 4 different types, materials, and colors of curtains in the laboratory scenario. We used ``Go through the curtain and watch out the chair." as the textual instruction. Given an arbitrary target behind the curtain, no feasible path could be planned without using our framework because the curtain separated the experimental scenario into two parts. With our framework, an action-aware costmap based on the textual instruction could be built and there would be a feasible path. The robot could pass through the curtain and reach the given target location following the path, as shown in Fig.~\ref{lab_exp} (a-d). 

Then we conducted the experiment of going through the curtain in a medical scenario at MRC. It contains a ``hospital bed", ``medical curtain", a ``warning sign", and a ``medical trolley". We used ``Deliver the medicine to me. By the way, you can go through the curtain and watch out the warning sign." as the textual instruction. As shown in Fig.~\ref{mrc_exp}, the robot avoided the warning sign, then passed through the curtain successfully, and finally reached the desired location. 

\subsubsection{Go through grass}
We prepared two types of artificial grass: setaria viridis and phragmites australis. We mixed two kinds of grass and used ``Go through the grass and watch out the chair." as the textual instruction. Similar to the curtain experiments, a feasible path could be planned and the robot went through the grass using our framework ( Fig.~\ref{lab_exp} (e)).

\section{CONCLUSION AND FUTURE WORK}
In the paper, we have presented a large-model-based interactive navigation framework, mapping textual instructions to feasible paths for navigating robots in the environment with traversable objects. To obtain a feasible path, we propose to construct an action-aware costmap by segmenting the LiDAR point clouds into traversable and untraversable parts, according to the bounding boxes of the landmarks and action-aware attributes from the large model module (LLM+VLM). Extensive experiments have shown the effectiveness and generalization of the proposed framework in different environments including a laboratory scenario and a medical scenario at MRC.

In the future, we plan to use a 3D LiDAR to replace the current 2D one to restore the 3D information of the environment for better sensing. To achieve more precise and robust action-aware segmentation of sensor data, we will deploy a state-of-the-art segmentation model (e.g., Segment Anything Model~\cite{kirillov2023segment}) to our LM module to produce high-quality object masks for improving the capability of segmenting sensor data (e.g., LiDAR point clouds) to the background and objects. Besides, we will upgrade the legged robot to a legged manipulator (e.g.,~\cite{10260538}), allowing us to manipulate traversable objects and thus enhancing interactive navigation.









\bibliographystyle{unsrt} 
\bibliography{bibliography.bib}

\begin{thebibliography}{10}

\bibitem{wellhausen2020safe}
Lorenz Wellhausen, Ren{\'e} Ranftl, and Marco Hutter.
\newblock Safe robot navigation via multi-modal anomaly detection.
\newblock {\em IEEE Robotics and Automation Letters}, 5(2):1326--1333, 2020.

\bibitem{vulcano2022safe}
Veronica Vulcano, Spyridon~G Tarantos, Paolo Ferrari, and Giuseppe Oriolo.
\newblock Safe robot navigation in a crowd combining nmpc and control barrier
  functions.
\newblock In {\em 2022 IEEE 61st Conference on Decision and Control (CDC)},
  pages 3321--3328. IEEE, 2022.

\bibitem{guan2022ga}
Tianrui Guan, Divya Kothandaraman, Rohan Chandra, Adarsh~Jagan Sathyamoorthy,
  Kasun Weerakoon, and Dinesh Manocha.
\newblock Ga-nav: Efficient terrain segmentation for robot navigation in
  unstructured outdoor environments.
\newblock {\em IEEE Robotics and Automation Letters}, 7(3):8138--8145, 2022.

\bibitem{liu2019obstacle}
Xuan Liu, Kashif~Nazar Khan, Qamar Farooq, Yunhong Hao, and Muhammad~Shoaib
  Arshad.
\newblock Obstacle avoidance through gesture recognition: Business advancement
  potential in robot navigation socio-technology.
\newblock {\em Robotica}, 37(10):1663--1676, 2019.

\bibitem{naseer2013followme}
Tayyab Naseer, J{\"u}rgen Sturm, and Daniel Cremers.
\newblock Followme: Person following and gesture recognition with a
  quadrocopter.
\newblock In {\em 2013 IEEE/RSJ International Conference on Intelligent Robots
  and Systems}, pages 624--630. IEEE, 2013.

\bibitem{ali2012real}
Muaammar Hadi~Kuzman Ali, M~Asyraf Azman, Zool~Hilmi Ismail, et~al.
\newblock Real-time hand gestures system for mobile robots control.
\newblock {\em Procedia Engineering}, 41:798--804, 2012.

\bibitem{stotko2019vr}
Patrick Stotko, Stefan Krumpen, Max Schwarz, Christian Lenz, Sven Behnke,
  Reinhard Klein, and Michael Weinmann.
\newblock A vr system for immersive teleoperation and live exploration with a
  mobile robot.
\newblock In {\em 2019 IEEE/RSJ International Conference on Intelligent Robots
  and Systems (IROS)}, pages 3630--3637. IEEE, 2019.

\bibitem{olatunji2022levels}
Samuel~A Olatunji, Andre Potenza, Andrey Kiselev, Tal Oron-Gilad, Amy Loutfi,
  and Yael Edan.
\newblock Levels of automation for a mobile robot teleoperated by a caregiver.
\newblock {\em ACM Transactions on Human-Robot Interaction (THRI)},
  11(2):1--21, 2022.

\bibitem{hu2019safe}
Zhe Hu, Jia Pan, Tingxiang Fan, Ruigang Yang, and Dinesh Manocha.
\newblock Safe navigation with human instructions in complex scenes.
\newblock {\em IEEE Robotics and Automation Letters}, 4(2):753--760, 2019.

\bibitem{brown2020language}
Tom Brown, Benjamin Mann, Nick Ryder, Melanie Subbiah, Jared~D Kaplan, Prafulla
  Dhariwal, Arvind Neelakantan, Pranav Shyam, Girish Sastry, Amanda Askell,
  et~al.
\newblock Language models are few-shot learners.
\newblock {\em Advances in neural information processing systems},
  33:1877--1901, 2020.

\bibitem{taylor2022galactica}
Ross Taylor, Marcin Kardas, Guillem Cucurull, Thomas Scialom, Anthony
  Hartshorn, Elvis Saravia, Andrew Poulton, Viktor Kerkez, and Robert Stojnic.
\newblock Galactica: A large language model for science.
\newblock {\em arXiv preprint arXiv:2211.09085}, 2022.

\bibitem{hoffmann2022training}
Jordan Hoffmann, Sebastian Borgeaud, Arthur Mensch, Elena Buchatskaya, Trevor
  Cai, Eliza Rutherford, Diego de~Las Casas, Lisa~Anne Hendricks, Johannes
  Welbl, Aidan Clark, et~al.
\newblock Training compute-optimal large language models.
\newblock {\em arXiv preprint arXiv:2203.15556}, 2022.

\bibitem{wei2022chain}
Jason Wei, Xuezhi Wang, Dale Schuurmans, Maarten Bosma, Fei Xia, Ed~Chi, Quoc~V
  Le, Denny Zhou, et~al.
\newblock Chain-of-thought prompting elicits reasoning in large language
  models.
\newblock {\em Advances in Neural Information Processing Systems},
  35:24824--24837, 2022.

\bibitem{shah2023lm}
Dhruv Shah, B{\l}a{\.z}ej Osi{\'n}ski, Sergey Levine, et~al.
\newblock Lm-nav: Robotic navigation with large pre-trained models of language,
  vision, and action.
\newblock In {\em Conference on Robot Learning}, pages 492--504. PMLR, 2023.

\bibitem{shah2023vint}
Dhruv Shah, Ajay Sridhar, Nitish Dashora, Kyle Stachowicz, Kevin Black, Noriaki
  Hirose, and Sergey Levine.
\newblock Vint: A foundation model for visual navigation.
\newblock {\em arXiv preprint arXiv:2306.14846}, 2023.

\bibitem{anderson2018vision}
Peter Anderson, Qi~Wu, Damien Teney, Jake Bruce, Mark Johnson, Niko
  S{\"u}nderhauf, Ian Reid, Stephen Gould, and Anton Van Den~Hengel.
\newblock Vision-and-language navigation: Interpreting visually-grounded
  navigation instructions in real environments.
\newblock In {\em Proceedings of the IEEE conference on computer vision and
  pattern recognition}, pages 3674--3683, 2018.

\bibitem{gasparino2022wayfast}
Mateus~V Gasparino, Arun~N Sivakumar, Yixiao Liu, Andres~EB Velasquez, Vitor~AH
  Higuti, John Rogers, Huy Tran, and Girish Chowdhary.
\newblock Wayfast: Navigation with predictive traversability in the field.
\newblock {\em IEEE Robotics and Automation Letters}, 7(4):10651--10658, 2022.

\bibitem{huang2023visual}
Chenguang Huang, Oier Mees, Andy Zeng, and Wolfram Burgard.
\newblock Visual language maps for robot navigation.
\newblock In {\em 2023 IEEE International Conference on Robotics and Automation
  (ICRA)}, pages 10608--10615. IEEE, 2023.

\bibitem{chen20232}
Peihao Chen, Xinyu Sun, Hongyan Zhi, Runhao Zeng, Thomas~H. Li, Gaowen Liu,
  Mingkui Tan, and Chuang Gan.
\newblock $a^2$nav: Action-aware zero-shot robot navigation by exploiting
  vision-and-language ability of foundation models, 2023.

\bibitem{chen2021pix2seq}
Ting Chen, Saurabh Saxena, Lala Li, David~J Fleet, and Geoffrey Hinton.
\newblock Pix2seq: A language modeling framework for object detection.
\newblock {\em arXiv preprint arXiv:2109.10852}, 2021.

\bibitem{du2022learning}
Yu~Du, Fangyun Wei, Zihe Zhang, Miaojing Shi, Yue Gao, and Guoqi Li.
\newblock Learning to prompt for open-vocabulary object detection with
  vision-language model.
\newblock In {\em Proceedings of the IEEE/CVF Conference on Computer Vision and
  Pattern Recognition}, pages 14084--14093, 2022.

\bibitem{li2022grounded}
Liunian~Harold Li, Pengchuan Zhang, Haotian Zhang, Jianwei Yang, Chunyuan Li,
  Yiwu Zhong, Lijuan Wang, Lu~Yuan, Lei Zhang, Jenq-Neng Hwang, et~al.
\newblock Grounded language-image pre-training.
\newblock In {\em Proceedings of the IEEE/CVF Conference on Computer Vision and
  Pattern Recognition}, pages 10965--10975, 2022.

\bibitem{liu2023grounding}
Shilong Liu, Zhaoyang Zeng, Tianhe Ren, Feng Li, Hao Zhang, Jie Yang, Chunyuan
  Li, Jianwei Yang, Hang Su, Jun Zhu, et~al.
\newblock Grounding dino: Marrying dino with grounded pre-training for open-set
  object detection.
\newblock {\em arXiv preprint arXiv:2303.05499}, 2023.

\bibitem{5477164}
Léonard Jaillet, Juan Cortés, and Thierry Siméon.
\newblock Sampling-based path planning on configuration-space costmaps.
\newblock {\em IEEE Transactions on Robotics}, 26(4):635--646, 2010.

\bibitem{5980048}
Jim Mainprice, E.~Akin~Sisbot, Léonard Jaillet, Juan Cortés, Rachid Alami,
  and Thierry Siméon.
\newblock Planning human-aware motions using a sampling-based costmap planner.
\newblock In {\em 2011 IEEE International Conference on Robotics and
  Automation}, pages 5012--5017, 2011.

\bibitem{lu2014layered}
David~V Lu, Dave Hershberger, and William~D Smart.
\newblock Layered costmaps for context-sensitive navigation.
\newblock In {\em 2014 IEEE/RSJ International Conference on Intelligent Robots
  and Systems}, pages 709--715. IEEE, 2014.

\bibitem{7324184}
Marina Kollmitz, Kaijen Hsiao, Johannes Gaa, and Wolfram Burgard.
\newblock Time dependent planning on a layered social cost map for human-aware
  robot navigation.
\newblock In {\em 2015 European Conference on Mobile Robots (ECMR)}, pages
  1--6, 2015.

\bibitem{7140063}
Yoichi Morales, Atsushi Watanabe, Florent Ferreri, Jani Even, Tetsushi Ikeda,
  Kazuhiro Shinozawa, Takahiro Miyashita, and Norihiro Hagita.
\newblock Including human factors for planning comfortable paths.
\newblock In {\em 2015 IEEE International Conference on Robotics and Automation
  (ICRA)}, pages 6153--6159, 2015.

\bibitem{7745154}
Omar A.~Islas Ramírez, Harmish Khambhaita, Raja Chatila, Mohamed Chetouani,
  and Rachid Alami.
\newblock Robots learning how and where to approach people.
\newblock In {\em 2016 25th IEEE International Symposium on Robot and Human
  Interactive Communication (RO-MAN)}, pages 347--353, 2016.

\bibitem{hart1968formal}
Peter~E Hart, Nils~J Nilsson, and Bertram Raphael.
\newblock A formal basis for the heuristic determination of minimum cost paths.
\newblock {\em IEEE transactions on Systems Science and Cybernetics},
  4(2):100--107, 1968.

\bibitem{hartley_zisserman_2004}
Richard Hartley and Andrew Zisserman.
\newblock {\em Multiple View Geometry in Computer Vision}.
\newblock Cambridge University Press, 2 edition, 2004.

\bibitem{legged_control}
Qiayuan Liao and Shang Yangxing.
\newblock legged\_control.
\newblock \url{https://github.com/qiayuanliao/legged_control}, 2022.

\bibitem{koenig2004design}
Nathan Koenig and Andrew Howard.
\newblock Design and use paradigms for gazebo, an open-source multi-robot
  simulator.
\newblock In {\em 2004 IEEE/RSJ international conference on intelligent robots
  and systems (IROS)(IEEE Cat. No. 04CH37566)}, volume~3, pages 2149--2154.
  IEEE, 2004.

\bibitem{hess2016real}
Wolfgang Hess, Damon Kohler, Holger Rapp, and Daniel Andor.
\newblock Real-time loop closure in 2d lidar slam.
\newblock In {\em 2016 IEEE international conference on robotics and automation
  (ICRA)}, pages 1271--1278. IEEE, 2016.

\bibitem{kirillov2023segment}
Alexander Kirillov, Eric Mintun, Nikhila Ravi, Hanzi Mao, Chloe Rolland, Laura
  Gustafson, Tete Xiao, Spencer Whitehead, Alexander~C Berg, Wan-Yen Lo, et~al.
\newblock Segment anything.
\newblock {\em arXiv preprint arXiv:2304.02643}, 2023.

\bibitem{10260538}
Xiangyu Chu, Shengzhi Wang, Minjian Feng, Jiaxi Zheng, Yuxuan Zhao, Jing Huang,
  and K.~W. Samuel~Au.
\newblock Model-free large-scale cloth spreading with mobile manipulation:
  Initial feasibility study.
\newblock In {\em 2023 IEEE 19th International Conference on Automation Science
  and Engineering (CASE)}, pages 1--6, 2023.

\end{thebibliography}

\end{document}